\documentclass[10pt,journal,compsoc]{IEEEtran}
\ifCLASSOPTIONcompsoc
  \usepackage[nocompress]{cite}
\else
  \usepackage{cite}
\fi
\usepackage{hyperref}       
\usepackage{url}            
\usepackage{booktabs}       
\usepackage{amsfonts}       
\usepackage{nicefrac}       
\usepackage{microtype}      

\usepackage{graphicx}
\usepackage{graphics}
\usepackage{multirow}
\usepackage{multicol}
\usepackage{algorithmic,algorithm}
\usepackage{amsmath,amssymb}
\usepackage{dsfont}
\usepackage{diagbox}

\newcommand{\eg}{\textit{e}.\textit{g}.}
\newcommand{\etal}{\textit{et al}.\ }
\DeclareMathOperator*{\argmax}{argmax}

\DeclareMathOperator*{\E}{\mathbb{E}}
\usepackage{enumitem}
\newtheorem{prop}{Proposition}
\newlist{Properties}{enumerate}{2}
\setlist[Properties]{label=Property \arabic*., font=\textbf, itemindent=*}
\usepackage{xcolor}

\ifCLASSINFOpdf
\else
\fi
\begin{document}
%
\title{Adversarially Robust Neural Architectures}
%
%
%
%

\author{Minjing~Dong,
        Yanxi~Li,
        Yunhe~Wang,
        and~Chang~Xu
\thanks{M. Dong, Y. Li and C. Xu are with the School of Computer Science, Faculty of Engineering, University of Sydney, 6 Cleveland Street, Darlington, NSW 2008, Australia. (e-mail: \{mdon0736@uni., yali0722@uni., c.xu@\}sydney.edu.au).}\\
\thanks{Yunhe Wang is with the Noah’s Ark Laboratory, Huawei Technologies Co., Ltd, HuaWei Building, No.3 Xinxi Road, ShangDi Information Industry Base, Hai-Dian District, Beijing 100085, P.R. China. (e-mail: yunhe.wang@huawei.com)}
}

%
%

\markboth{Journal of \LaTeX\ Class Files,~Vol.~14, No.~8, August~2015}%
{Shell \MakeLowercase{\textit{et al.}}: Bare Demo of IEEEtran.cls for Computer Society Journals}
%



\IEEEtitleabstractindextext{%
\begin{abstract}
Deep Neural Networks (DNNs) are vulnerable to adversarial attacks. Existing methods are devoted to developing various robust training strategies or regularizations to update the weights of the neural network.  But beyond the weights, the overall structure and information flow in the network are explicitly determined by the neural architecture, which remains unexplored.  This paper thus aims to improve the adversarial robustness of the network from the architecture perspective.
We explore the relationship among adversarial robustness, Lipschitz constant, and architecture parameters and show that an appropriate constraint on architecture parameters could reduce the Lipschitz constant to further improve the robustness. The importance of architecture parameters could vary from operation to operation or connection to connection. 
We approximate the Lipschitz constant of the entire network through a univariate log-normal distribution, whose mean and variance are related to architecture parameters. The confidence can be fulfilled through formulating a constraint on the distribution parameters based on the cumulative function.
Compared with adversarially trained neural architectures searched by various NAS algorithms as well as efficient human-designed models, our algorithm empirically achieves the best performance among all the models under various attacks on different datasets.
\end{abstract}

}

\maketitle

\IEEEdisplaynontitleabstractindextext

%
\IEEEpeerreviewmaketitle

\IEEEraisesectionheading{\section{Introduction}\label{sec:introduction}}
\IEEEPARstart{D}{eep} neural networks have shown remarkable performance in various applications, such as image classification \cite{Krizhevsky:2012:ICD:2999134.2999257,DBLP:journals/corr/SimonyanZ14a,DBLP:journals/corr/HeZR016}, object detection \cite{DBLP:journals/corr/HuangLW16a}, and machine translation \cite{DBLP:journals/corr/BahdanauCB14,chen2020distilling}. However, recent works \cite{DBLP:journals/corr/NguyenYC14,DBLP:journals/corr/Moosavi-Dezfooli15,DBLP:journals/corr/Moosavi-Dezfooli16,DBLP:journals/corr/abs-1708-06131,DBLP:journals/corr/abs-1812-03411} have shown that DNNs are vulnerable to adversarial samples that can fool the networks to make wrong predictions with only perturbations of the input data, which has caused security issues. To deal with the threat of adversarial samples, the majority of existing works focus on robust training which optimizes the weights of robust DNNs through feeding adversarial samples generated by attack approaches (\eg\ FGSM).
Although the trained networks show good robustness on various attacks, the architectures of these networks are fixed during optimization, which limits the adversarial robustness improvement. The efficient architectures designed by human experts, such as AlexNet and ResNet \cite{NIPS2012_4824,DBLP:journals/corr/HeZR016}, suggest that the DNN performance is subject to the architecture of network. Recent boosting NAS studies also emphasize the influence of architecture. Hence, we ask a simple question:

\emph{
Can the network be initialized with robust architecture to further obtain adversarial robustness?}

A recent study has shown that different architectures tend to have different levels of adversarial robustness \cite{guo2019nas}. Thus, the designing of robust neural architectures becomes essential for robustness improvement.
However, the problem remains since designing a robust neural architecture can be rather expensive due to the substantial time cost and human effort, and the direct relationship between adversarial robustness and architectures is still unexplored.

To reduce the cost of discovering superior robust neural architecture, we made use of NAS algorithms which automatically discover the ideal architectures within a predefined search space. Recently, remarkable progress has achieved in NAS, including RL-based approaches \cite{DBLP:journals/corr/ZophL16,DBLP:journals/corr/BakerGNR16,DBLP:journals/corr/abs-1709-07417} and gradient-based approaches \cite{DBLP:journals/corr/abs-1806-09055,inproceedings_gdas,DBLP:journals/corr/abs-1907-05737,tang2020semi}. In particular, DARTS \cite{DBLP:journals/corr/abs-1806-09055} introduced a differentiable method for architecture optimization through a continuous relaxation on discrete search space through forming a weighted sum of operations instead of discrete architecture selection, which significantly reduced the searching budget.

Although the NAS framework provided an efficient way to automatically discover the superior neural architectures with customized objective, the standard adversarial training required massive cost in generating the adversarial examples, which significantly decreased the search efficiency. Thus, we tried to dismiss the inner maximum of adversarial training to further accelerate the optimization of architecture through involving the Lipschitz constraint by exploring the influence of Lipschitz constant on adversarial robustness and how the architecture parameters impact the Lipschitz constant. In this paper, we proposed to explore the relationship between adversarial robustness and the architecture of network through establishing their connections to Lipschitz constant under NAS framework.

Furthermore, the instability of differentiable NAS algorithm has been explored by previous work \cite{Zela2020Understanding}. Existing differentiable NAS algorithms used to utilize architecture parameters for sampling superior architectures, where all the elements of architecture parameters are ``equally treated" for selection without exploring their discrepancies. For example, two nodes in the same cell may have different levels of freedom of selecting operation, however, they were only assigned with trainable parameters and applied with $argmax$ for selection after searching, which significantly reduced the reliability of sampled architecture and raises a demand for confidence learning of architecture parameters. Thus, we proposed to sample architecture parameters from trainable distributions instead of initializing them directly.

Our proposed algorithm Adversarially Robust Neural Architecture Search with Confidence Learning (RACL) starts from the approximation of Lipschitz constant of entire neural network under NAS framework, where we derive the relationship between Lipschitz constant and architecture parameters. We further propose to sample architecture parameters from log-normal distributions. With the usage of the properties of log-normal distribution, we show that the Lipschitz constant of entire network can be approximated with another log-normal distribution with mean and variance related to architecture parameters so that a constraint can be formulated in a form of cumulative function to achieve Lipschitz constraint on the architecture. Our algorithm achieves an efficient robust architecture search and RACL empirically achieves superior adversarial robustness compared with other NAS algorithms as well as state-of-the-art models through a series of experiments under different settings.

\section{Related Work}
To situate the current work, we review some relevant literature that covers adversarial attacks, defend methods and neural architecture search.

\subsection{Adversarial Attacks}
Szegedy \etal first revealed the adversarial samples, which demonstrated that neural networks are vulnerable to adversarial attacks \cite{2013arXiv1312.6199S}. Given a fixed input, through utilizing the model gradient w.r.t the input, a perturbation which wildly changes the predicted output can be easily found. Vast techniques have been introduced to generate powerful adversarial samples in a efficient way. Adversarial attacks are generally divided into two groups, the white-box case \cite{goodfellow2014explaining,DBLP:journals/corr/Moosavi-Dezfooli16,DBLP:journals/corr/KurakinGB16a,DBLP:journals/corr/abs-1801-02610} and black-box case \cite{DBLP:journals/corr/PapernotMGJCS16,Chen_2017,DBLP:journals/corr/abs-1712-07113,DBLP:journals/corr/abs-1905-06635}. The white-box cases enable attacks have full access to the network. Goodfellow \etal introduced an efficient attack method FGSM through one-step gradient-based attack on the input \cite{DBLP:journals/corr/Moosavi-Dezfooli16}. Kurakin \etal \cite{DBLP:journals/corr/KurakinGB16} first proposed an iterative attack methods I-FGSM instead of one-step gradient-based attack to obtain much more powerful attacks. Dong \etal proposed to integrate momentum into the I-FGSM for more stable updating and boost the transferability of generated adversarial samples. Mardry introduced a strong attack, Projected Gradient Descent (PGD), which is now widely used in robustness learning \cite{aleks2017deep}. PGD attack utilized the local first-order information of the network to achieve high attack success rates. Universal adversarial perturbations have also been studied by previous work \cite{8423654}.
On the contrary, the black-box attacks where the model architecture and parameters are not accessible are relatively weak. However, the black-box attacks fit the actual circumstances better and thus have received many attentions. Madry \etal explored the transferability phenomenon of adversarial attacks which showed that the generated adversarial samples can also achieve relatively high attack success rates on another network \cite{aleks2017deep}. Besides transfer-based black-box attacks, some query-based attacks were introduced where adversaries can only query the outputs of the models \cite{DBLP:journals/corr/abs-1712-07113,ilyas2019prior}. Yan \etal \cite{DBLP:journals/corr/abs-1906-04392} proposed to bridge the gap between transfer-based and query-based attacks to achieve more efficient black-box attacks.
Besides the attacks on the classification tasks, adversarial attacks have been applied to other tasks, such as detection and segmentation \cite{metzen2017universal,DBLP:journals/corr/XieWZZXY17}.

\subsection{Defence Mechanisms against Attacks}
Due to the exponential growth of attack approaches, more attention have been paid to defence methods recently which tackled the vulnerability of neural networks through improving the adversarial robustness. There are various defence mechanisms proposed by previous work.
Gradient Masking methods hided the gradient information to confound the adversaries \cite{DBLP:journals/corr/PapernotM17,DBLP:journals/corr/abs-1710-10766}, however, they cannot defend attacks based on approximate gradient \cite{DBLP:journals/corr/abs-1802-00420}.
Adversarial Example Detection is another stream which aims at discovering the adversarial examples and rejects them \cite{DBLP:journals/corr/HendrycksG16b,DBLP:journals/corr/GrosseMP0M17}. Feinman \etal proposed to randomize the classifier with Dropout to identify the adversarial examples based on the prediction variance \cite{feinman2017detecting}.
The main stream of defence Mechanisms is the robust optimization which further optimizes the network to achieve adversarial robustness. Adversarial training is naturally introduced to defend attacks through feeding adversarial examples into the training stage to form a min-max game where the inner maximum generates adversarial samples to maximize the classification loss and outer minimum optimizes model parameters to minimize the loss. Different attack strategies have been applied to generate adversarial examples for adversarial training, such as PGD attack \cite{aleks2017deep} and FGSM attack \cite{goodfellow2014explaining}. Besides the standard adversarial training, different variants have been proposed. Miyato \etal proposed virtual adversarial training which defines the adversarial direction without label information \cite{8417973}. Shafahi \etal introduced an efficient adversarial training which recycled the gradient information \cite{NIPS2019_8597}. Zhang \etal proposed to decompose the prediction error into the classification and boundary error and provided a tight upper bound \cite{DBLP:journals/corr/abs-1901-08573}. Pang \etal \cite{DBLP:journals/corr/abs-1901-08846} introduced adaptive diversity promoting (ADP) which improved the adversarial robustness of ensemble models.
Some regularization methods have been introduced to defend against attacks. \cite{cisse2017parseval,weng2018evaluating} proposed to constrain the Lipschitz constant of network to improve the adversarial robustness. Mustafa \etal introduced an effective constraint which forced the features for each class to lie inside a convex polytope and separated from those of other classes \cite{9025211}.

\subsection{Neural Architecture Search}
Although vast approaches have been proposed to defend against adversarial samples, most of them focused on optimizing the weights based on different strategies, and the impact of architecture has been ignored. Recently, neural architecture search has received increasing attention due to its superior performance. Early NAS approaches heavily relied on macro searching which directly searches the entire network \cite{DBLP:journals/corr/abs-1708-05344,DBLP:journals/corr/ZophL16}. For efficiency, more NAS approaches have applied micro search space where the cell is searched instead of the entire network, and the cells are stacked in series to compose the whole network \cite{DBLP:journals/corr/abs-1802-03268,DBLP:journals/corr/ZophVSL17}. Yang \etal proposed an efficient continuous evolutionary approach on the supernet where all the architectures share the parameters, which boosted the searching efficiency \cite{Yang_2020_CVPR}. Recently, the differentiable searching algorithm DARTS has been introduced to boost the searching speed through a relaxation on search space to form a supernet with operation mixture to achieve differentiable architecture searching \cite{DBLP:journals/corr/abs-1806-09055}. Dong \etal introduced a differentiable sampler over the supernet with Gumbel-Softmax which improved the searching efficiency \cite{inproceedings_gdas}. Xu \etal proposed to apply channel sampling for searching acceleration and add edge normalization to stabilize the searching phase \cite{DBLP:journals/corr/abs-1907-05737}. Recently, NAS has been applied to different areas, including adversarial robustness. Guo \etal empirically demonstrated that different architectures had different levels of robustness and proposed feature flow guided search to discover the robust neural architectures \cite{guo2019nas}. Chen \etal proposed ABanditNAS with improved conventional bandit algorithm to search the architectures under enlarged search space to better defend against adversarial attacks \cite{chen2020antibandit}. \textcolor{black}{One concern of NAS for adversarial robustness is the computational cost since both adverarial training and supernet optimization can be time-consuming. Kotyan \etal \cite{DBLP:journals/corr/abs-1906-11667} investigated the potential robust architecture in a broader search space including the concatenation and connections between dense and convolution layers, and demonstrated that there exist robust architectures which achieve inherent accuracy on adversarial examples. Different from \cite{DBLP:journals/corr/abs-1906-11667}, our objective aims at discovering the robust architecture in the current popular search space benchmark \cite{DBLP:journals/corr/abs-1806-09055,inproceedings_gdas} without expensive adversarial training via investigating the connection between Lipschitz constraint and adversarial robustness of architecture.}


\section{Methodology}
In this section, we introduce the proposed robust architecture search with confidence learning (RACL) algorithm. Different from existing defence methods which only focus on weights optimization, we lay emphasis on the influence of architecture on adversarial robustness through exploring the relationship among robustness, architecture, and Lipschitz constant. Confidence learning is further involved to form a Lipschitz constraint on architecture parameters. With the proposed algorithm, the searched architectures can have stronger defensive power against adversarial examples.
\subsection{Preliminary}
Given the input $x\in \mathbb{R}^D$ and annotated label vector $y \in \mathbb{R}^M$ where $M$ is the total number of classes, the neural network \textcolor{black}{$\mathcal{H}$} maps perturbed input $\tilde{x}=x+\delta$ to a label vector \textcolor{black}{$\hat{y} = \mathcal{H}(\tilde{x};W, \mathcal{A})$}. The network architecture is represented by $\mathcal{A}$, and its filter weight is denoted as $W$. The objective of adversarial attacks is to find the perturbed input $\tilde{x}$ which leads to wrong predictions through maximizing the classification loss as
\begin{equation}
\begin{aligned}
     \color{black}\tilde{x} = \argmax_{\tilde{x}:\|\tilde{x}-x\|_p \leqslant \epsilon} \mathcal{L}_{CE}(\mathcal{H}(\tilde{x};W, \mathcal{A}),y),
\end{aligned}
\end{equation}
where $\mathcal{L}_{CE}(\hat{y},y) = -\sum^M_{i=1}y^{(i)}log(\hat{y}^{(i)})$, and the perturbation is constrained by its $l_p$-norm.
Various powerful attacks have been proposed and shown high attack success rates, such as Fast Gradient Sign Method (FGSM) \cite{2013arXiv1312.6199S} and Projected Gradient Descent (PGD) \cite{aleks2017deep}. To defend against these attacks, regularizing the weight matrix of each layer to form a Lipschitz-constrained network has been proven to be beneficial for the adversarial robustness \cite{cisse2017parseval,weng2018evaluating}. 

Let \textcolor{black}{$\mathcal{F}=\mathcal{L} \circ \mathcal{H}$} be the mapping from the input to the classification loss, and the difference of loss after an adversarial attack can be bounded as
\begin{equation} \label{eq:Lipschitz_constant}
   \|\mathcal{F}(x+\delta,y;W,\mathcal{A}) - \mathcal{F}(x,y;W,\mathcal{A})\| \leqslant \lambda_{\mathcal{F}} \|\delta\|,
\end{equation}
where $\lambda_{\mathcal{F}}$ is the Lipschitz constant of function $\mathcal{F}$ with respect to $\|.\|_p$. Together with $\|\delta\|_p \leqslant \epsilon$, the generalization error with perturbed input can be bounded as
\begin{equation} \label{eq:generalization_error}
\begin{aligned}
    \E_{x \backsim \mathcal{D}} [\mathcal{F}(\tilde{x})] & \leqslant \E_{x \backsim \mathcal{D}} [\mathcal{F}(x)] +  \E_{x \backsim \mathcal{D}} [\max_{\|\tilde{x}-x\| \leqslant \epsilon}|\mathcal{F}(\tilde{x})-\mathcal{F}(x)|] \\
    & \leqslant \E_{x \backsim \mathcal{D}} [\mathcal{F}(x)] + \lambda_{\mathcal{F}} \cdot \epsilon, \\
\end{aligned}
\end{equation}
which suggests that neural networks can defend against adversarial examples with a smaller Lipschitz constant. \textcolor{black}{Although it is difficult to derive the precise Lipschitz constant $\lambda_{\mathcal{F}}$ given a network, we can impose constraints on both the lower bound and upper bound of Lipschitz constant, which are denoted as $\underline{\lambda_{\mathcal{F}}}$ and $\overline{\lambda_{\mathcal{F}}}$ respectively. Thus, an adversarial robust formulation of neural architectures can be written as}
\begin{equation}\label{eq:Lipschitz_objective}
    \color{black}{\min_{\mathcal{A},W} \E [\mathcal{F}(x,y;W,\mathcal{A})] \ s.t.\  \underline{\lambda_{\mathcal{F}}^*} \leqslant \lambda_{\mathcal{F}} \leqslant \overline{\lambda_{\mathcal{F}}^*},}
\end{equation}
\textcolor{black}{where $\underline{\lambda_{\mathcal{F}}^*}$ and $\overline{\lambda_{\mathcal{F}}^*}$ are the optimal lower and upper bounds of Lipschitz constant.}
Existing works often consider a fixed network architecture $\mathcal{A}$ in Eq. (\ref{eq:Lipschitz_objective}), and focus on optimizing network weight for improved robustness, where the influence of architecture is ignored. Recent studies highlight the importance of architecture. Liu \etal conducts thorough experiments to empirically demonstrate that the better trade-offs of some pruning techniques mainly come from the architecture itself \cite{DBLP:journals/corr/abs-1810-05270}. Boosting NAS algorithms involve optimization of architecture to obtain better performance with small model size \cite{DBLP:journals/corr/abs-1806-09055,DBLP:journals/corr/abs-1907-05737}. We are therefore motivated to investigate the influence of neural architecture on adversarial robustness. 
\begin{figure*}[t]
\begin{center}
\includegraphics[width=1.0\linewidth]{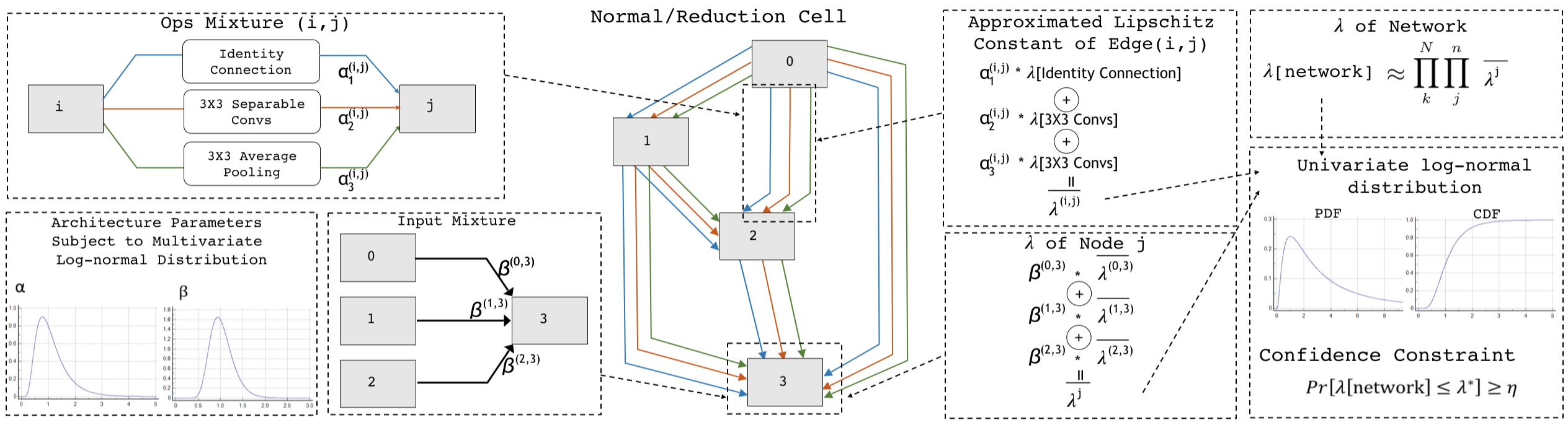}
\end{center}
\caption{An overview of proposed robust neural architecture search with confidence learning algorithm. Each node in the cell is computed with operations mixture under architecture parameters $\alpha$ for weighting operations and $\beta$ for weighting inputs where $\alpha$ and $\beta$ are sampled from multivariate log-normal distributions. Meanwhile, the Lipschitz constant of each edge and cell induce to univariate log-normal distributions. The Lipschitz constraint is formulated from cumulative distribution function.}
\label{fig:architecture}
\end{figure*}
\subsection{Lipschitz Constraints in Neural Architecture}

The discrete architecture $\mathcal{A}$ is determined by both connections and operations, which creates a huge search space. Differentiable Architecture Search algorithms provide an efficient solution through the continuous relaxation of the architecture representation \cite{DBLP:journals/corr/abs-1806-09055,DBLP:journals/corr/abs-1712-00559,DBLP:journals/corr/abs-1802-01548}. Within the differentiable NAS framework, we decompose the entire neural network into cells. Each cell $I$ is a directed acyclic graph (DAG) consisting of an ordered sequence of $n$ nodes, where each node denotes a latent representation that is transformed from two previous latent representations and each edge $(i,j)$ denotes an operation $o$ from a pre-defined search space $\mathcal{O}$ which transforms $I^{(i)}$. Following \cite{DBLP:journals/corr/abs-1907-05737}, the architecture parameters $\alpha$ which weighs operations, and $\beta$ which weighs input flows are introduced to form an operation mixture with weighted inputs. The intermediate node is computed as 
\begin{equation} \label{eq:node_computation}
    I^{(j)} = \sum_{i<j}\beta^{(i,j)} \sum_{o\in \mathcal{O}}  {\alpha}_o^{(i,j)} \cdot o(I^{(i)}),
\end{equation}
where $I^{(0)}$ and $I^{(1)}$ are fixed as inputs nodes during the searching phase and the last node is formed by channel-wise concatenating of previous intermediate nodes $I = \cup^{n-1}_{i=2} I^{(i)}$ as the output of cell. 

The entire neural network is constructed through two different types of cells including the normal cell, where all the operations have strides of 1, and the reduction cell, where the operations connected to the two inputs have strides of 2. With normal and reduction cells stacked in series, the entire neural network can be formed as \textcolor{black}{$\mathcal{H} = I_1 \circ I_2 \circ \dots \circ I_N \circ \mathcal{C}$}, where $N$ denotes the number of cells and $\mathcal{C}$ denotes the classifier. Following \cite{DBLP:journals/corr/abs-1907-05737}, after the searching phase, the operation $o$ with the maximum $\beta^{(i,j)} \alpha_o^{(i,j)}$ for each edge $(i,j)$ is selected and the connection of each node $j$ to its two precedents $i<j$ with maximum $\beta^{(i,j)} \alpha_o^{(i,j)}$ is selected so that a discrete superior architecture can be sampled from the supernet.

We now explore the relationship between architecture parameters $\alpha$, $\beta$ and Lipschitz constant of the network. Since the entire neural network is constructed by stacking cells in series as $[I_1, I_2, ..., I_N]$, Eq. \ref{eq:Lipschitz_constant} can be further decomposed as as
\begin{equation} \label{eq:Lipschitz_network}
    \begin{aligned}
        \|\mathcal{F}(\tilde{x}) - \mathcal{F}(x)\| & \leqslant 
        \color{black}\lambda_{l} \|\mathcal{H}(\tilde{x}) - \mathcal{H}(x)\| \\
        & \leqslant \lambda_{l} \lambda_{\mathcal{C}} \|I_{N}(\tilde{x}) - I_{N}(x)\| \\
        & \leqslant \lambda_{l} \lambda_{\mathcal{C}} \lambda (I_N) \|I_{N-1}(\tilde{x}) - I_{N-1}(x)\|, \\
    \end{aligned}
\end{equation}
where $\lambda_l$, $\lambda_{\mathcal{C}}$ and $\lambda (I_N)$ denote the Lipschitz constants of the loss function, classifier, and cell $I_N$ respectively. By rewriting $\|I_{N-1}(\tilde{x}) - I_{N-1}(x)\|$ in a format of its previous cells till the input of cell becomes the image for $I_{1}$ and considering $\|I_{1}(\tilde{x}) - I_{1}(x)\| \leqslant \lambda (I_1) \|\tilde{x}-x\| = \lambda (I_1) \| \delta \|$, Eq. \ref{eq:Lipschitz_network} can be unfolded recursively and rewritten as
\begin{equation} \label{eq:Lipschitz_network_2}
   \|\mathcal{F}(\tilde{x}) - \mathcal{F}(x)\| \leqslant \lambda_{\mathcal{F}} \|\delta\| \leqslant \|\delta\| \lambda_{l} \lambda_{\mathcal{C}} \prod_k^N \lambda(I_k).
\end{equation}
It is obvious that the adversarial robustness can be bounded by the Lipschitz constants of cells. Eq. \ref{eq:Lipschitz_network_2} also suggests that the impact of perturbation grows exponentially with the number of cells, which further highlights the influence of cell designing. 

As $\lambda_{l}$ and $\lambda_{\mathcal{C}}$ in Eq. \ref{eq:Lipschitz_network_2} are not related to the architecture,  we next focus on the discussion on $\lambda(I_k)$. Based on the operation mixture defined in Eq. \ref{eq:node_computation}, the variation of node $I^{(j)}_k$ under perturbation can be written in a format of that in previous node $I^{(j)}_k$.  For simplicity of notation, we omit the subscript $k$ and for each node and we have
\begin{equation} \label{eq:intermediate_node}
\begin{aligned}
    & \|I^{(j)}(\tilde{x})-I^{(j)}(x)\| \leqslant \sum_{i<j} \beta^{(i,j)} \lambda^{(i,j)} \|I^{(i)}(\tilde{x}) - I^{(i)}(x)\|, \\
    & \ s.t.\ \lambda^{(i,j)} \leqslant \sum_{o\in \mathcal{O}} \alpha_o^{(i,j)} \lambda_{o},\\
\end{aligned}
\end{equation}
where $\lambda^{(i,j)}$ denotes the Lipschitz constant of transformation from node i to j and $\lambda_o$ denotes the Lipschitz constant of operation $o$. Similarly, we can unfold Eq. \ref{eq:intermediate_node} recursively for entire cell by rewriting $\|I^{(i)}(\tilde{x}) - I^{(i)}(x)\|$ in a format of its previous node, and have
\begin{equation} \label{eq:Lipschitz_cell}
            \lambda (I^{(j)}) \leqslant \sum_{i<j} \beta^{(i,j)} \sum_{o\in \mathcal{O}} \alpha_o^{(i,j)} \lambda_{o}.
\end{equation}
Through substituting $\lambda(I_k)$ in Eq. \ref{eq:Lipschitz_network_2} by the one in Eq. \ref{eq:Lipschitz_cell} and taking $\lambda_{l}$ and $\lambda_{\mathcal{C}}$ as a unified constant $C$, the lipschitz constant $ \lambda_{\mathcal{F}}$ is bounded by the product of the Lipschitz constant of intermediate nodes as
\begin{equation} \label{eq:lipschitz_full_eq}
\begin{aligned}
    \lambda_{\mathcal{F}} & \leqslant C \prod_k^N \lambda (I_k) \leqslant C \prod_k^N \prod^n_j \lambda (I^{(j)}) \\
    & \leqslant C \prod_k^N \prod_j^n \sum_{i<j} \beta^{(i,j)} \sum_{o\in \mathcal{O}} \alpha_o^{(i,j)} \lambda_{o}. \\
\end{aligned}
\end{equation}
According to the definition, the Lipschitz constant of operations without convolutional layers can be summarized as follows, (1). average pooling: $S^{-0.5}$ where $S$ denotes the stride of pooling layer, (2). max pooling: $1$,  (3). identity connection: $1$, (4). Zeroize: $0$. For the rest operations including depth-wise separate conv and dilated depth-wise separate conv, we focus on the $L_2$ bounded perturbations and according to the definition of spectral norm, the Lipschitz constant of these operations is the spectral norm of its weight matrix where $\lambda^{o}_2 = \|W^{o}\|_2$, which also is the maximum singular value of $W$, marked as $\Lambda_1$. However, directly computing $\Lambda_1$ is not practical through gradient descent. To achieve a differentiable optimization on Lipschitz constant, we make use of the power iteration method which can be applied for an efficient approximation of $\Lambda_1$ \cite{yoshida2017spectral}. Note that although the perturbation is $L_2$ bounded, the robustness against $L_\infty$ can be also achieved, as stated by \cite{DBLP:journals/corr/abs-1802-07896}.

\begin{figure}[t]
\begin{center}
\includegraphics[width=1.0\linewidth]{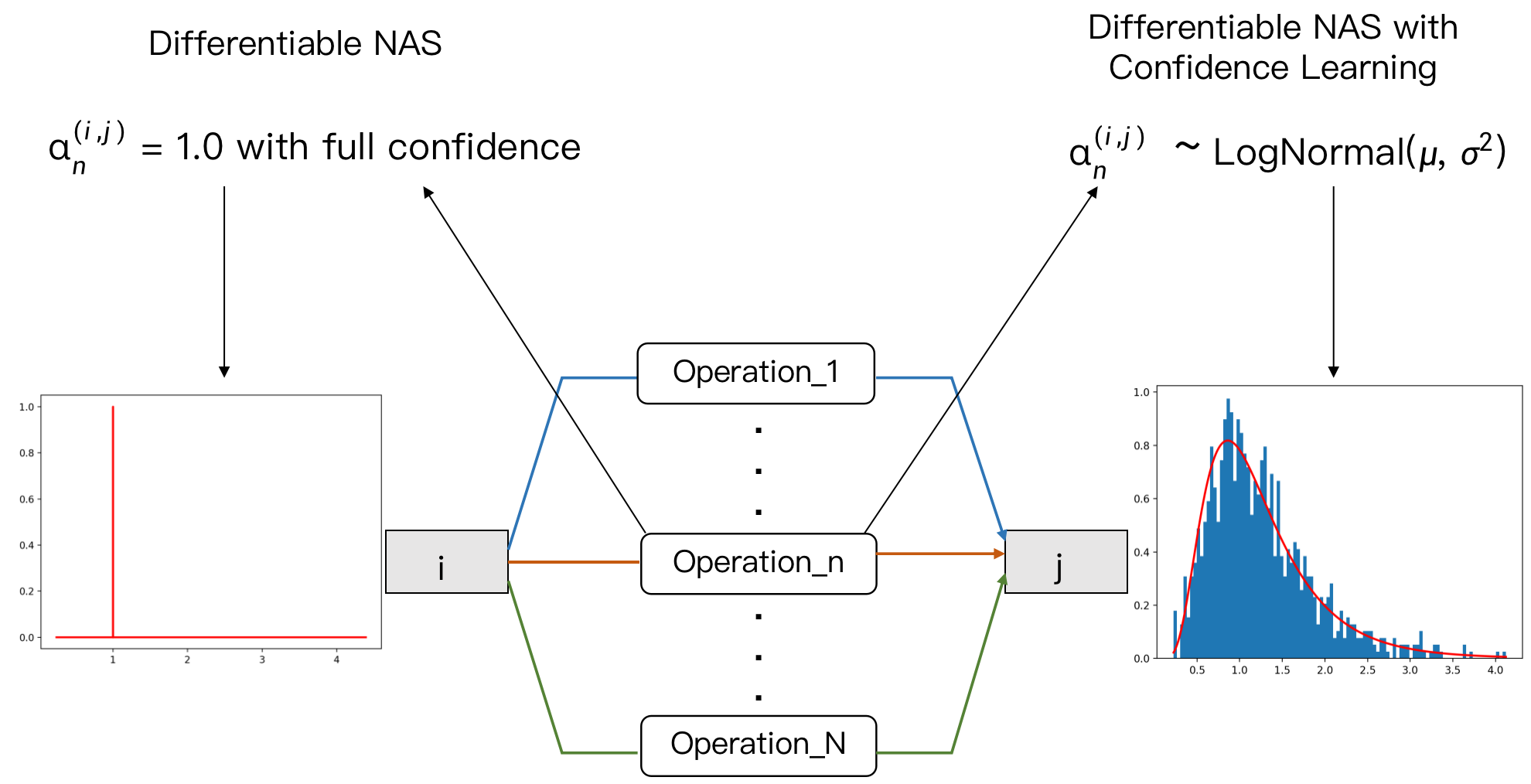}
\end{center}
\caption{An illustration of the difference between previous differentiable NAS and ours with confidence learning.}
\label{fig:Confidence_fig}
\end{figure}

\subsection{Confident Architecture Sampling}
The architecture is determined by parameters $\alpha$ and $\beta$, which further influences the Lipschitz constants of the network, as shown in Eq. \ref{eq:lipschitz_full_eq}. Existing NAS algorithms used to initialize them as trainable parameters without in-depth analysis. However, these weightings on operations or connections naturally could have different levels of importance and freedom. \textit{E}.\textit{g}., the connection of the first node and one of the intermediate nodes may have different levels of freedom for the final selection, but they are simply assigned values with the same confidence for optimization and sampled with the maximum value in NAS framework. Thus, previous architecture parameters can hardly fulfill this requirement. Instead, we propose to explore the confidence on the architecture parameters by regarding them as variables sampled from distributions during architecture search. An illustration of the advantage of confidence learning is shown in Fig \ref{fig:Confidence_fig}. For each architecture parameters, previous differentiable NAS algorithms made use of trained value with full confidence as shown in the left part, while our algorithm enables parameters to exploit their confidence as shown in the right part. Intuitively, the architecture optimization is highly uncertain due to the large search space. But existing NAS absolutely trusts all architecture parameters without discriminating the confidence on them. Taking $\alpha$ and $\beta$ from the perspective of distributions, the variances will be optimized to indicate confidence in the values. RACL tends to exploit the operations of higher confidence and explore more potential good paths by investigating operations of lower confidence. The overall searching space can thus be well explored and exploited.


For distributions, a naive selection can be a multivariate normal distribution. However, according to the Lipschitz constant form in Eq. \ref{eq:lipschitz_full_eq}, the sampled values from this distribution need to be positive since $\lambda_\mathcal{F}$ is always positive and negative values from distribution will make the constraint cease to be effective. Thus, we turn to log-normal distribution $\mathcal{LN}$ since it guarantees positive sampled values. Note that a random variable is log-normally distributed $\mathcal{LN}(\mu, \Sigma)$ if the logarithm of it is normally distributed $\mathcal{N}(\mu, \Sigma)$. For simplicity, the following mean and variance denote those of the logarithm. Most importantly, there are several nice properties, including the weighted sum of multiple independent $\mathcal{LN}_{1,...,n}$ can be approximated with another $\mathcal{LN}$ and the product of multiple independent $\mathcal{LN}_{1,...,n}$ induces to $\mathcal{LN}$ with parameters $\mu$ and $\Sigma$ of the sum of those in $\mathcal{LN}_{1,...,n}$. Thus we propose to sample $\alpha$ from multivariate log-normal distributions, denoted as $\mathcal{LN}(\mu^\alpha, \Sigma^\alpha)$, with mean $\mu^\alpha \in \mathbb{R}^d$ and covariance matrix $\Sigma^\alpha \in \mathbb{R}^{d \times d}$ with diagonal standard deviation $\sigma^\alpha \in \mathbb{R}^d$ where $d$ denotes the dimension of $\alpha$. Similarly, we sample $\beta$ from $\mathcal{LN}(\mu^\beta, \Sigma^\beta)$.

Back to the Lipschitz constant, the multivariate log-normal distribution over $\alpha$ induces a univariate log-normal distribution over the upper boundary of Lipschitz constant of edge based on the operation mixture $\sum_{o\in \mathcal{O}} \alpha_o^{(i,j)} \lambda_{o}$ since it can be treated as the weighted sum of multiple log-normal distributed variables. Note that $\lambda_{o}$ is treated as constant here since the weights are fixed when optimizing architecture parameters. The is proposed Lipschitz confidence constraint is shown in Fig .\ref{fig:architecture}. Although there is no closed-form expression of its probability density function, the distribution can be approximated using the properties of log-normal distribution as:
\begin{Properties}
  \item If a log-normal variable $X \backsim \mathcal{LN}(\mu, \sigma^2)$ is multiplied by a constant $a$, $aX \backsim \mathcal{LN}(\mu + ln(a), \sigma^2)$. \label{properties:1}
  \item If multiple independent log-normal variables, denoted as $X_1, X_2, ..., X_n$, are multiplied, $X_1 \cdot X_2 \cdots X_n \backsim \mathcal{LN}(\sum_{i=1}^n\mu_i, \sum_{i=1}^n\sigma_i^2)$ \label{properties:2}
\end{Properties}
Following \cite{6771360}, the sum of log-normal distributions can be approximated by another log-normal distribution as below:
\begin{prop} \label{prop:1}
If multiple independent log-normal variables, denoted as $X_1, X_2, ..., X_n$, are added, the sum $Z=\sum^n_{i=1} X_i$ can be approximated by another log-normal distribution $\mathcal{LN}(\mu_Z, \sigma_Z^2)$ with variance $\sigma_Z^2=ln[\frac{\sum e^{(2(\mu_i)+{\sigma_i^2})} (e^{(\sigma_i)^2}-1) }{(\sum e^{(\mu_+\sigma_i^2/2)})^2}+1])$ and mean $\mu_Z = ln[\sum e^{(\mu_i+\sigma_i^2/2)}] - \frac{\sigma_Z}{2}$.
\end{prop}
Thus, in Eq. \ref{eq:intermediate_node}, the distribution of $\sum_{o\in \mathcal{O}} \alpha_o^{(i,j)} \lambda_{o}$ can be treated as a weighted sum of multiple independent log-normal distributions and can be approximated with these properties and Proposition \ref{prop:1}. Similarly, in Eq. \ref{eq:Lipschitz_cell}, $\sum_{i<j} \beta^{(i,j)} \sum_{o\in \mathcal{O}} \alpha_o^{(i,j)} \lambda_{o}$ can be treated as the sum of multiple products of two independent log-normal distributions which can also be approximated. Based on the properties of log-normal distribution, the upper boundary of Lipschitz constant of entire network can be approximated accordingly,
\begin{equation} \label{eq:lambda_pdf}
\begin{aligned}
    & \overline{\lambda^{(i,j)}} = \sum_{o\in \mathcal{O}} \alpha_o^{(i,j)} \lambda_{o}\\
    & \backsim \mathcal{LN}(ln[\sum_o e^{({\mu_o^\alpha}'+(\sigma^\alpha_o)^2/2)}] - \frac{\sigma^2_{I^{(i,j)}}}{2}, \sigma^2_{I^{(i,j)}}), \\
    & \sigma^2_{I^{(i,j)}} = ln[\frac{\sum_o e^{(2({\mu_o^\alpha}')+{(\sigma^\alpha_o)^2})} (e^{(\sigma^\alpha_o)^2}-1) }{(\sum_o e^{({\mu_o^\alpha}')+(\sigma^\alpha_o)^2/2)})^2}+1]), \\
    & {\mu_o^\alpha}' = \mu_o^\alpha+ln(\lambda_{o})\\
\end{aligned}
\end{equation}
where $\overline{\lambda^{(i,j)}}$ denotes the upper boundary of $\lambda^{(i,j)}$. For simplicity, we denote the mean for $\overline{\lambda^{(i,j)}}$ as $\mu_{I^{(i,j)}}$ and variance as $\sigma^2_{I^{(i,j)}}$. Similarly, we sample $\beta$ from a multivariate log-normal distribution $\mathcal{N}(\mu^\beta, \Sigma^\beta)$. For variable $\beta^{(i,j)} \overline{\lambda^{(i,j)}}$, it can be treated as the product of two log-normal distributions, which also follows a log-normal distribution whose mean is the sum of means of two distributions and variance as well. Thus, to generalize the distribution over edge $\overline{\lambda^{(i,j)}}$ to the one over intermediate node $\overline{\lambda^{(j)}}$, we replace $o$ with $j$, $\mu^\alpha_o + ln(\lambda_o)$ with $\mu^\beta_{(i,j)}+\mu_{I^{(i,j)}}$, and $(\sigma^\alpha_o)^2$ with $(\sigma^\beta_{(i,j)})^2+\sigma^2_{I^{(i,j)}}$ in Eq \ref{eq:lambda_pdf} and obtain the log-normal distribution of $\overline{\lambda^{(j)}} = \sum_{i<j} \beta^{(i,j)} \overline{\lambda^{(i,j)}}$ as
\begin{equation}
\begin{aligned}
    & \overline{\lambda^{(j)}} = \sum_{i<j} \beta^{(i,j)} \overline{\lambda^{(i,j)}} \\
    & \backsim \mathcal{LN}(ln[\sum_o e^{({\mu^\beta_{(i,j)}}' +([{\sigma^\beta_{(i,j)}}’]/2)}] - \frac{\sigma^2_{I^{(j)}}}{2}, \sigma^2_{I^{(j)}}), \\
    & \sigma^2_{I^{(j)}} = ln[\frac{\sum_j e^{(2({\mu^\beta_{(i,j)}}' )+{[{\sigma^\beta_{(i,j)}}’]})} (e^{[{\sigma^\beta_{(i,j)}}’]}-1) }{(\sum_o e^{({\mu^\beta_{(i,j)}}' +[({\sigma^\beta_{(i,j)}}’]/2)})^2}+1]) \\
    & {\mu^\beta_{(i,j)}}' = \mu^\beta_{(i,j)}+\mu_{I^{(i,j)}},\\
    & {\sigma^\beta_{(i,j)}}’ = (\sigma^\beta_{(i,j)})^2+\sigma^2_{I^{(i,j)}}\\
\end{aligned}
\end{equation}
with mean and variance which are denoted as $\mu_{I^{(j)}}$ and $\sigma^2_{I^{(j)}}$. According to Eq. \ref{eq:lipschitz_full_eq}, $\lambda_{\mathcal{F}}$ is bounded by the product of $\overline{\lambda^{(j)}}$. Thus, $\overline{\lambda_{\mathcal{F}}}$ follows the log-normal distribution with mean $\mu = In(C)+\sum^N_k\sum^n_j\mu_{I^{(j)}}$ and variance $\sigma^2 = \sum^N_k\sum^n_j\sigma^2_{I^{(j)}}$. We introduce a confidence hyperparameter $\eta \in [0,1]$ to enable confidence learning with such an constraint as
\begin{equation} \label{eq:confidence_1}
\begin{aligned}
    & {Pr}_{\alpha,\beta} [\overline{\lambda_{\mathcal{F}}} \leqslant \overline{\lambda_{\mathcal{F}}^*}] \\
    & = {Pr}_{\alpha,\beta} [C \prod_k^N \prod_j^n \sum_{i<j} \beta^{(i,j)} \sum_{o\in \mathcal{O}} \alpha_o^{(i,j)} \lambda_{o} \leqslant \overline{\lambda_{\mathcal{F}}^*}] \geqslant \eta, \\
\end{aligned}
\end{equation}
where $\overline{\lambda_{\mathcal{F}}^*}$ is the desired Lipschitz constant upper bound of $\mathcal{F}$, Note that in Eq. \ref{eq:confidence_1}, the variance of $\overline{\lambda_{\mathcal{F}}}$ is reduced to satisfy the inequality, which strengths the confidence on the approximation of Lipschitz constant of $\lambda_\mathcal{F}$, compared with the one in Eq. \ref{eq:lipschitz_full_eq} without confidence learning. To obtain a convex constraint in $\mu$ and $\Sigma$, we reformulate Eq. \ref{eq:confidence_1} through the format of cumulative function as
\begin{equation}
\begin{aligned}
    Pr[\overline{\lambda_{\mathcal{F}}} \leqslant \overline{\lambda_{\mathcal{F}}^*}] & = Pr[\frac{ln(\overline{\lambda_{\mathcal{F}}})-\mu}{\sigma} \leqslant \frac{ln(\overline{\lambda_{\mathcal{F}}^*})-\mu}{\sigma}]\\
    & = \Phi(\frac{ln(\overline{\lambda_{\mathcal{F}}^*})-\mu}{\sigma}),\\
\end{aligned}
\end{equation}
where $\Phi$ denotes the cumulative function of the normal distribution since $\frac{ln(\overline{\lambda_{\mathcal{F}}})-\mu}{\sigma}$ is a random variable following the normal distribution. Thus, we establish direct relationship among $\mu$, $\sigma$ and $\eta$ as
\begin{equation} \label{eq:upper_bound_constraint}
    \frac{ln(\overline{\lambda_{\mathcal{F}}^*})-\mu}{\sigma} \geqslant \Phi^{-1}(\eta).
\end{equation}
Through omitting the square root on $\sigma$, we achieve a convex constraint. \textcolor{black}{Besides the upper bound of Lipschitz constant, we propose to minimize the lower bound $\underline{\lambda_{\mathcal{F}}}$ together to better control $\lambda_{\mathcal{F}}$. Taking the advantage of the fact that $\|\nabla \mathcal{F}(x,y;W,\mathcal{A})\| \leq \lambda_{\mathcal{F}}$, we simply take $\underline{\lambda_{\mathcal{F}}} = \|\nabla \mathcal{F}(x,y;W,\mathcal{A})\|$.}
\textcolor{black}{Together with the constraint in Eq. \ref{eq:upper_bound_constraint}, we reformulate the optimization objective in Eq. \ref{eq:Lipschitz_objective} as}
\begin{equation} \label{eq:constraint_objective}
\begin{aligned}
        & \color{black}\min_{\mu^\alpha, \Sigma^\alpha, \mu^\beta, \Sigma^\beta,W} \mathcal{L}_{CE}(\mathcal{F}(x;W, \mathcal{A}),y) + \|\nabla \mathcal{F}(x,y;W,\mathcal{A})\|, \\
        & \  s.t.\ ln(\overline{\lambda_{\mathcal{F}}^*}) - \mu \geqslant \Phi^{-1}(\eta) \sigma^2, \\
        & \mathcal{A} \backsim \mathcal{LN}(\mu^\alpha, \Sigma^\alpha), \mathcal{LN}(\mu^\beta, \Sigma^\beta).\\
\end{aligned}
\end{equation}
Intuitively, the constraint in Eq. \ref{eq:constraint_objective} reveals the influence of $\sigma$ on sampling architecture parameters. As $\sigma$ increases, the value of $\mu$ decreases to satisfy the inequality where the corresponding $\mu^\alpha$ and $\mu^\beta$ become $0$ for relatively large $\sigma$, which implies that the operations or connections are unlikely to be sampled when its corresponding confidence is low. Thus, the architecture can be sampled based on its confidence in the Lipschitz constraint.
We apply the ADMM optimization framework to solve this constrained optimization through incorporating the constraint to form a minimax problem so that Eq. \ref{eq:constraint_objective} can be rewritten as
\begin{equation}
    \begin{aligned}
        & \color{black}\min_{\mu^\alpha,\Sigma^\alpha,\mu^\beta,\Sigma^\beta}\max_{\theta} \mathcal{L}_{CE} + \underline{\lambda_{\mathcal{F}}} + \theta (c(\mu,\Sigma)) + \frac{\rho}{2}\|c(\mu,\Sigma)\|^2_F, \\
        & \color{black} c(\mu,\Sigma) = \mu + \Phi^{-1}(\eta)\sigma^2 - ln(\overline{\lambda_{\mathcal{F}}^*}),\\
    \end{aligned}
\end{equation}
where $\theta$ is the dual variable and $\rho$ is positive number predefined in ADMM. The first step is to update $\mu$ while fixing other variables and the second step is to update $\sigma$ while fixing other variables as
\begin{equation} \label{eq:mu_sigma_update}
\begin{aligned}
    & \color{black}\mu_{t+1} \leftarrow \mu_t - \gamma \triangledown_\mu [\mathcal{L}_{CE} + \underline{\lambda_{\mathcal{F}}} + \theta (c(\mu,\Sigma_t)) + \frac{\rho}{2}\|c(\mu,\Sigma_t)\|^2_F], \\
    & \color{black}\sigma_{t+1} \leftarrow \sigma_t - \gamma \triangledown_\sigma [\mathcal{L}_{CE} + \underline{\lambda_{\mathcal{F}}} + \theta (c(\mu_t,\Sigma)) + \frac{\rho}{2}\|c(\mu_t,\Sigma)\|^2_F], \\
\end{aligned}
\end{equation}
where $\mu^\alpha,\Sigma^\alpha,\mu^\beta,\Sigma^\beta$ are updated through back-propagation. The dual variable $\theta$ is updated with learning rate of $\rho$ as
\begin{equation} \label{eq:dual_update}
    \theta_{t+1} \leftarrow \theta_t + \rho \cdot c(\mu_t,\Sigma_t)
\end{equation}
The entire robust neural architecture search with confidence learning algorithm, denoted as RACL, is shown in Alg. \ref{alg:Robust_alg}. With proposed algorithm, we impose confidence learning on the values of architecture parameters $\alpha$ and $\beta$, which strengthens the confidence of robust architecture sampling.
\begin{algorithm}[t]
  \caption{Robust Neural Architecture Search with High Confidence Algorithm}
  \label{alg:Robust_alg}
\begin{algorithmic}
  \STATE {\bf{Input:}} The training set is split into $\mathcal{D}_T$ and $\mathcal{D}_V$; Batch size n; Hyperparameter $\lambda^*$, $\rho$, $\eta$;
  \STATE Initialize multivariate log-normal distributions $\mathcal{LN}(\mu^\alpha, \Sigma^\alpha)$ and $\mathcal{LN}(\mu^\beta, \Sigma^\beta)$; Initialize \textcolor{black}{$\mathcal{H}$} with $W$;
  \WHILE{not converge}
  \STATE Sample $\alpha$ and $\beta$ from $\mathcal{LN}(\mu^\alpha, \Sigma^\alpha)$ and $\mathcal{LN}(\mu^\beta, \Sigma^\beta)$ based on reparameterization trick
  \STATE Sample batch of data $\{(x_i, y_i)\}_{i=1}^n$ from $\mathcal{D}_T$
  \STATE \textcolor{black}{Optimize $W$ with $\sum_{i=1}^n\mathcal{L}_{CE}(x_i,y_i;W,\alpha,\beta)+\underline{\lambda_{\mathcal{F}}}$}
  \STATE Sample batch of data $\{(x_i, y_i)\}_{i=1}^n$ from $\mathcal{D}_V$
  \STATE Optimize $\mu^\alpha, \Sigma^\alpha, \mu^\beta, \Sigma^\beta $ with ADMM framework
  \STATE \textcolor{black}{$\mu_{t+1} \leftarrow \mu_t - \gamma \triangledown_\mu [\mathcal{L}_{CE} + \underline{\lambda_{\mathcal{F}}} + \theta (c(\mu,\Sigma_t)) + \frac{\rho}{2}\|c(\mu,\Sigma_t)\|^2_F]$}
  \STATE \textcolor{black}{$\sigma_{t+1} \leftarrow \sigma_t - \gamma \triangledown_\sigma [\mathcal{L}_{CE} + \underline{\lambda_{\mathcal{F}}} + \theta (c(\mu_t,\Sigma)) + \frac{\rho}{2}\|c(\mu_t,\Sigma)\|^2_F]$}
  \STATE Optimize $\theta$ with $\theta_{t+1} \leftarrow \theta_t + \rho \cdot c(\mu_t,\Sigma_t)$
  \ENDWHILE
  \STATE Sample the normal and reduction cell based on sampled $\alpha$ and $\beta$
  \STATE Retrain the searched architecture from scratch on training set
\end{algorithmic}
\end{algorithm}

\begin{table*}[t]
\caption{\textcolor{black}{Evaluation of RACL adversarial robustness on CIFAR-10, CIFAR-100, and Tiny-ImageNet compared with various NAS algorithms under white-box attacks. PGD$^{20}$ denotes PGD attack with 20 iterations. Best results in bold.}} \label{table:white_box_NAS}
\begin{center}
\scalebox{1.0}{
\color{black}\begin{tabular}{c|cc|ccccccc}
\hline
Dataset & Model & Params & Natural & FGSM & MIM & PGD$^{20}$ & PGD$^{100}$ & CW & AutoAttack\\ \hline \hline
\multirow{5}{*}{CIFAR-10} 
& AmoebaNet    & 3.2M & 82.28\% & 59.12\% & 57.26\% & 53.69\% & 53.34\% & 78.63\% & 47.88\%\\
& NASNet       & 3.8M & \bf84.37\% & 61.38\% & 58.72\% & 53.35\% & 52.84\% & 80.69\% & 48.19\%\\
& DARTS        & 3.3M & 80.65\% & 59.63\% & 57.55\% & 54.04\% & 53.73\% & 77.01\% & 48.13\%\\
& PC-DARTS     & 3.6M & 84.32\% & 61.08\% & 58.10\% & 53.01\% & 52.36\% & 80.54\% & 47.95\%\\
\cline{2-10}
& RACL(ours)   & 3.3M & 84.04\% & \bf62.55\% & \bf60.00\% & \bf55.68\% & \bf55.32\% & \bf80.90\% & \bf50.07\%\\
\hline
\hline
\multirow{5}{*}{CIFAR-100} 
& AmoebaNet    & 3.2M & 56.51\% & 32.67\% & 31.44\% & 29.70\% & 29.66\% & 49.03\% & 25.26\%\\
& NASNet       & 3.8M & 57.97\% & 31.54\% & 30.14\% & 28.58\% & 28.44\% & 49.64\% & 24.42\%\\
& DARTS        & 3.3M & \bf58.67\% & 32.71\% & 31.14\% & 29.21\% & 29.11\% & 50.32\% & 24.30\%\\
& PC-DARTS     & 3.6M & 57.20\% & 31.85\% & 30.46\% & 28.62\% & 28.50\% & 49.40\% & 24.10\%\\
\cline{2-10}
& RACL(ours)   & 3.3M & 57.83\% & \bf33.89\% & \bf32.41\% & \bf30.41\% & \bf30.15\% & \bf52.56\% & \bf25.55\%\\
\hline
\hline
\multirow{5}{*}{Tiny-ImageNet} 
& AmoebaNet    & 3.2M & 47.84\% & 31.44\% & 30.57\% & 30.12\% & 30.09\% & 42.56\% & -\\
& NASNet       & 3.8M & 47.85\% & 30.76\% & 29.80\% & 29.47\% & 29.44\% & 41.93\% & -\\
& DARTS        & 3.3M & 48.20\% & 31.38\% & 30.71\% & 30.30\% & 30.25\% & 42.23\% & -\\
& PC-DARTS     & 3.6M & 47.24\% & 30.04\% & 29.18\% & 28.55\% & 28.53\% & 40.91\% & -\\
\cline{2-10}
& RACL(ours)   & 3.3M & \bf48.86\% & \bf31.98\% & \bf31.12\% & \bf30.63\% & \bf30.63\% & \bf42.99\% & -\\
\hline
\end{tabular}
}
\end{center}
\end{table*}

\section{Experiments}
In this section, we conduct a series of experiments to empirically demonstrate the effectiveness of proposed RACL algorithm. We retrain the searched neural architecture and compare it with various neural architectures searched by NAS algorithms as well as state-of-the-art network architectures.
We show that under various adversarial attack settings, the robust neural architectures searched by RACL always achieve better robustness than other baselines.
\begin{figure}[t]
\begin{center}
\includegraphics[width=1.0\linewidth]{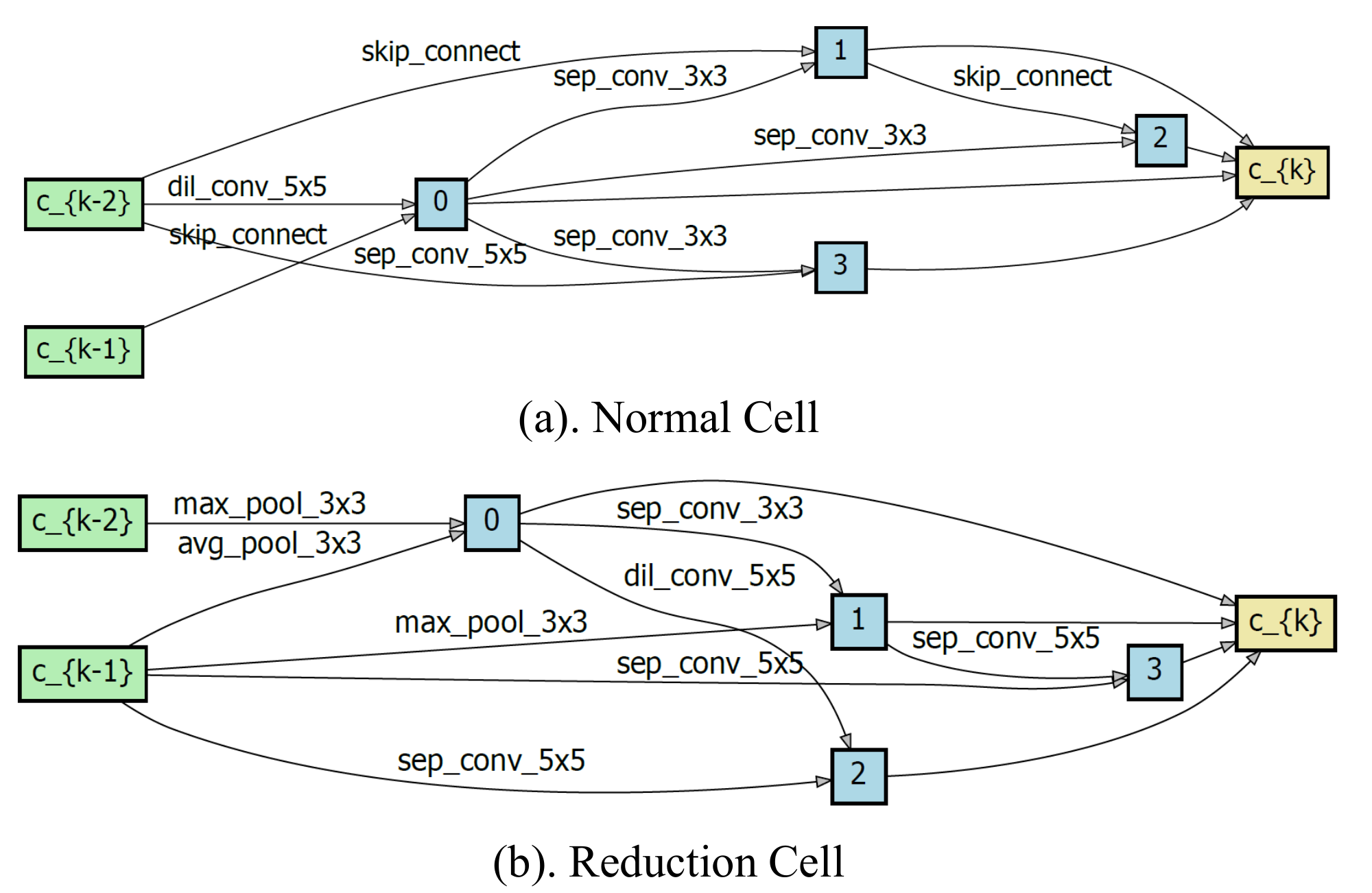}
\end{center}
\caption{\textcolor{black}{The visualization of normal and reduction cell searched by RACL are shown in (a) and (b).}}
\label{fig:Cell}
\end{figure}
\subsection{Experimental Setup}
\noindent\textbf{Neural Architecture Search Setup}\quad 
Following previous works \cite{DBLP:journals/corr/abs-1806-09055, DBLP:journals/corr/abs-1907-05737}, we search the robust neural architectures on CIFAR-10 dataset which contains 50K training images and 10K validation images over 10 classes. During the searching phase, the training set is divided into two parts with equal sizes for architecture and weight optimization respectively. The search space includes 8 candidates: $3\times3$ and $5\times5$ separable convolutions, $3\times3$ and $5\times5$ dilated separable convolutions, $3\times3$ max pooling, $3\times3$ average pooling, skip connection, and zero, as suggested by previous works \cite{DBLP:journals/corr/abs-1806-09055, DBLP:journals/corr/abs-1907-05737}. The supernet is constructed by stacking 8 cells including 6 normal cells and 2 reduction cells, each of which contains 6 nodes.  For the training settings, we follow the setups of PC-DARTS \cite{DBLP:journals/corr/abs-1806-09055}. The searching phase takes 50 epochs with a batch size of $128$. We use SGD with momentum. The initial learning rate is $0.1$ with a momentum of $0.9$ and a weight decay is $3 \times 10^{-4}$ to update the supernet weights. Architecture parameters were updated with Adam with a learning rate of $6 \times 10^{-4}$ and a weight decay of $1 \times 10^{-3}$. The searching time of RACL takes 0.5 GPU days.

\noindent\textbf{Datasets and Retrain Details}\quad
We extensively evaluate the proposed algorithm on three datasets including CIFAR-10, CIFAR-100, and \textcolor{black}{Tiny-ImageNet}, which are widely compared by other previous work. The searched superior neural architecture is sampled based on the proposed sampling strategy. \textcolor{black}{Following the setting in NAS \cite{DBLP:journals/corr/abs-1806-09055,DBLP:journals/corr/abs-1907-05737}, we stack searched cells to form a 20-layer network and retrained it with the entire training set.}
\textcolor{black}{For the evaluation stage, we adopt the popular adversarial training framework to retrain all the baselines. We train the network from scratch for 100 epochs with a batch size of 128 on the entire training set. We use SGD optimizer with an initial learning rate of $0.1$, momentum of $0.9$, and a weight decay of $2 \times 10^{-4}$. The norm gradient clipping is set to $5$. Following \cite{Wang2020Improving}, the hyperparameter which balances the adversarial loss and KL divergence is set to $6$. Since we focus on the impact of architecture on adversarial robustness, we train the searched architectures as well as state-of-the-art network architectures with the same adversarial training setting. Through training these architectures in the same adversarial manner, we conduct a fair comparison among different architectures and demonstrate how they improve or constrain the adversarial robustness. For CIFAR-10 and CIFAR-100, we use adversarial training with the total perturbation size $\epsilon = 8/255$. The maximum number of attack iterations is set to $10$ with a step size of $2/255$. For Tiny-ImageNet, we set $\epsilon = 4/255$. The maximum number of attack iterations is set to $6$ with a step size of $2/255$. An illustration of the searched normal cell and reduction cell is shown in Figure \ref{fig:Cell}, more analysis on searched robust neural architectures will be covered in Sec. \ref{sec: variance}}
\subsection{Against White-box Attacks}
\textcolor{black}{To evaluate the superiority of proposed RACL, we compare the searched cells with SOTA NAS algorithms, including DARTS \cite{DBLP:journals/corr/abs-1806-09055}, PC-DARTS \cite{DBLP:journals/corr/abs-1907-05737}, NASNet \cite{DBLP:journals/corr/ZophVSL17}, AmoebaNet \cite{DBLP:journals/corr/abs-1802-01548}. We also compare RACL with SOTA human-designed network architectures, such as ResNet and DenseNet \cite{DBLP:journals/corr/HeZR016,DBLP:journals/corr/HuangLW16a}. Furthermore, some NAS algorithms targeting the adversarial robustness are also included for comparison, including RobNet \cite{guo2019nas} and ABanditNAS \cite{chen2020antibandit}. Moreover, we also compare our results with other defence mechanisms, including Stochastic Weight Averaging (SWA) \cite{chen2021robust} and Instance Adaptive Adversarial training (IAAT) \cite{DBLP:journals/corr/abs-1910-08051}. For robustness evaluation, we choose various popular powerful attacks including Fast Gradient Sign Method (FGSM) \cite{2013arXiv1312.6199S}, Momentum Iterative Method (MIM) \cite{DBLP:journals/corr/abs-1710-06081}, Projected Gradient Descent (PGD) \cite{aleks2017deep}, CW attack \cite{carlini2017towards}, and Auto Attack \cite{DBLP:journals/corr/abs-2003-01690}. Consistent with previous adversarial literature \cite{aleks2017deep,DBLP:journals/corr/abs-1901-08573}, the perturbation is considered under $l_\infty$ norm with the total perturbation size of $8/255$ on CIFAR-10/100 and $4/255$ on Tiny-ImageNet. For CW attack, the steps are set to $1000$ with a learning rate of $0.01$.}

\begin{table}[t]
	\caption{\textcolor{black}{Evaluation of RACL adversarial robustness on CIFAR-10 compared with other human-designed architectures and other robust NAS algorithms under white-box attacks. Best results in bold.}}
	\label{table:robust_nas}
	\centering
	\scalebox{0.92}{
		\color{black}\begin{tabular}{@{}c@{}|@{}c@{}|ccccc}
			\hline
			Model               & Params & Natural & FGSM & PGD$^{20}$ & PGD$^{100}$ & MIM\\ \hline \hline
			ResNet-18           & 11.17M & 78.38\% & 49.81\% & 45.60\% & 45.10\% & 45.23\%\\
			ResNet-50           & 23.52M & 79.15\% & 51.46\% & 45.84\% & 45.35\% & 45.53\%\\
			WRN-28-10           & 36.48M & 86.43\% & 53.57\% & 47.10\% & 46.90\% & 47.04\%\\
			DenseNet-121        & 6.95M  & 82.72\% & 54.14\% & 47.93\% & 47.46\% & 48.19\%\\
			\hline
			ABanditNAS          & 5.19M  & \bf90.64\% & - & 50.51\% & - & 54.19\%\\
			RobNet-S            & 4.41M  & 78.05\% & 53.93\% & 48.32\% & 48.07\% & 48.98\%\\
			RobNet-M            & 5.56M  & 78.33\% & 54.55\% & 49.13\% & 48.96\% & 49.34\%\\
			RobNet-L        & 6.89M  & 78.57\% & 54.98\% & 49.44\% & 49.24\% & 49.92\%\\
			RobNet-free         & 5.49M  & 82.79\% & 58.38\% & 52.74\% & 52.57\% & 52.95\%\\ \hline
			RACL(ours)          & 3.34M  & 84.04\% & \bf{62.55\%} & \bf{55.68\%} & \bf{55.32\%} & \bf{60.00\%}\\
			\hline
	\end{tabular}}
\end{table}

\noindent\textbf{\textcolor{black}{Evaluation on CIFAR-10, CIFAR-100, and Tiny-ImageNet}}\quad 
\textcolor{black}{
Although adversarial training is a strong defence method, the impact of architecture is always ignored. In this experiment, we demonstrate that constructing networks via the neural architectures searched by RACL can further improve the robustness after adversarial training.
For a fair comparison, we retrain the searched cells using PGD adversarial training for all the models to evaluate the robustness of RACL on the main benchmark of defence mechanisms. The number of PGD attack iterations is set to $20$ and $100$ with a step size of $2/255$, as suggested by \cite{guo2019nas}. The detailed evaluation results are shown in Table \ref{table:white_box_NAS}. The best result for each column is highlighted in bold.
As shown in Table \ref{table:white_box_NAS}, RACL achieves better adversarial accuracy than other state-of-the-art neural architectures on all the datasets. For example, compared with our baseline PC-DARTS on CIFAR-10, though both RACL and PC-DARTS achieve similar clean accuracy and model size, their performance with adversarial training varies differently. RACL achieves an accuracy of $62.55\%$ under FGSM attack, with $1.47\%$ improvement ($61.08\% \to 62.55\%$) over that of PC-DARTS, and $2.96\%$ improvement ($52.36\% \to 55.32\%$) over that of PC-DARTS under PGD$^{100}$ attack. Furthermore, RACL achieves the best robust accuracy than other baselines under different attacks on CIFAR-100 and Tiny-ImageNet. For example, RACL achieves an accuracy of $52.26\%$ under Auto Attack, with $1.25\%$ improvement ($24.30\% \to 25.55\%$) over that of DARTS on CIFAR-100. Similarly, RACL achieves an accuracy of $42.99\%$ under CW attack, with $1.06\%$ improvement ($41.93\% \to 42.99\%$) over that of NASNet on Tiny-ImageNet. 
We empirically show that RACL consistently achieves the best robust performance compared with other NAS algorithms with the same search space under various attacks, which indicates that the adversarial robustness can be further improved through imposing Lipschitz constraint on architecture parameters.
}

\noindent\textbf{\textcolor{black}{Comparison with Human-designed Architectures and Robust NAS Algorithms on CIFAR-10}}\quad
\textcolor{black}{
Besides the standard NAS algorithms, there exist some NAS algorithm targeting adversarial robustness as well as some popular human-designed architectures which are widely compared in adversarial robustness benchmarks. We include ResNet-18, ResNet-50, WideResNet-28-10, and DenseNet-121 for comparison. The results are shown in Table \ref{table:robust_nas}. Compared with these human-designed architectures, RACL shows obvious superiority of robust accuracy over all the baselines under various attacks with fewer parameters. In terms of robust NAS algorithms, RobNet applies robust architecture search algorithm to explore a RobNet family under different budgets \cite{guo2019nas}. Compared with RobNet-S, RobNet-M and RobNet-L, RACL consistently achieves the best performance with a large gap. Compared with RobNet-free which relaxes the cell-based constraint, RACL still achieves better results with fewer parameters. For example, RACL achieves an accuracy of $55.32\%$ under PGD$^{100}$ attack, with $2.75\%$ improvement ($52.57\% \to 55.32\%$) over that of RobNet-free. Compared with ABanditNAS which includes denoise operations in its search space, RACL outperforms it in adversarial accuracy. For example, RACL achieves an accuracy of $55.32\%$ under PGD$^{100}$ attack, with $5.81\%$ improvement ($54.19\% \to 60.00\%$) over that of ABanditNAS. Overall, RACL achieves superior trade-offs among parameters, clean accuracy, and adversarial accuracy, which highlights the effectiveness and efficiency of proposed RACL algorithm.
}

\begin{table}[t]
\caption{\textcolor{black}{Comparison with existing defence techniques under PGD attack on different datasets.}} \label{table:AT_comparison}
\begin{center}
\scalebox{0.98}{
\color{black}\begin{tabular}{@{}c@{}|@{}c@{}|c|c|c}
\hline
Attack & Defence & CIFAR-10 & CIFAR-100 & Tiny-ImageNet\\ 
\hline
\hline
\multirow{4}{*}{PGD$^{20}$} 
& FAT \cite{DBLP:journals/corr/abs-2002-11242}    
                    & 45.31\% & 27.38\% & 17.50\%\\
& SWA \cite{chen2021robust}       
                    & 52.14\% & 28.28\% & 21.84\%\\
& NADAR \cite{DBLP:journals/corr/abs-2108-06885}
                    & 53.43\% & 28.40\% & 21.14\%\\
& RACL(ours)        & \bf{55.68\%} & \bf{30.41\%} & \bf{30.63\%}\\
\hline
\multirow{4}{*}{PGD$^{100}$} &
IAAT \cite{DBLP:journals/corr/abs-1910-08051}
                    & 46.50\% & 24.22\% & - \\
& RobNet-L \cite{guo2019nas}  
                    & 49.24\% & 23.19\% & 19.90\% \\
& RobNet-free \cite{guo2019nas}   
                    & 52.57\% & 23.87\% & 20.87\% \\
& RACL(ours)                      
                    & \bf{55.32\%} & \bf{30.15\%} & \bf{21.48\%}\\
\hline
\end{tabular}
}
\end{center}
\end{table}
\noindent\textbf{Comparison with existing defence mechanisms}\quad \textcolor{black}{We argue that initializing a network with robust neural architecture can be regarded as an efficient defence method against adversarial samples. To illustrate how robust architecture improves the performance of adversarial training, we compare RACL with previously proposed defence mechanisms on different datasets, including CIFAR-10, CIFAR-100, and Tiny-ImageNet. 
The perturbation budget $\epsilon$ is set to $8/255$ for CIFAR-10 and CIFAR-100. For Tiny-ImageNet, we consider two perturbation budgets. Following \cite{DBLP:journals/corr/abs-2108-06885,chen2021robust}, the perturbation budget of PGD attack $\epsilon$ is set to $4/255$ with 20 iterations as PGD$^{20}$. We also consider another stronger attacking setting in Tiny-ImageNet, which follows \cite{guo2019nas}. The perturbation budget of PGD attack $\epsilon$ is set to $8/255$ with 100 iterations as PGD$^{100}$.
We include various defence mechanisms for comparison. RACL was also compared with FAT \cite{DBLP:journals/corr/abs-2002-11242} which aims at better trade-offs between natural accuracy and robustness, SWA \cite{chen2021robust} which introduces weight smoothing to tackle the overfitting issue in adversarial training, IAAT \cite{DBLP:journals/corr/abs-1910-08051} which enforces sample-specific perturbation margins for a better generalization, and NADAR \cite{DBLP:journals/corr/abs-2108-06885} which proposes to search the dilation network for adversarial robustness. The detailed results are shown in Table \ref{table:AT_comparison}. 
Compared with all the SOTA defence techniques, RACL consistently achieves the best performance in all the scenarios, which demonstrates the superiority of RACL as defence mechanism. Note that RACL can collaborate with other adversarial training algorithms to achieve potentially better performance.}
\begin{table}[t]
		\caption{\textcolor{black}{Evaluation of RACL adversarial robustness on CIFAR-10 under transfer-based black-box attack setting.}} \label{table:black_box_NAS}
	\begin{center}
		\scalebox{1.0}{
			\color{black}\begin{tabular}{c|ccc}
				\hline
				Model & FGSM & MIM & PGD$^{20}$\\ \hline \hline
				AmoebaNet & 81.92\% & 82.04\% & 82.61\%\\
				NasNet    & 82.32\% & 82.60\% & 83.21\%\\
				DARTS     & 78.76\% & 78.83\% & 79.29\%\\
				PC-DARTS  & 82.47\% & 82.65\% & 83.06\%\\ \hline
				RACL(ours)& \bf{82.78\%} & \bf{82.92\%} & \bf{83.45\%}\\
				\hline
			\end{tabular}
		}
	\end{center}
\end{table}
\subsection{Against Black-box Attacks}
\noindent\textbf{Transfer-based Black-box Attack Evaluation}\quad We next evaluate the robustness of RACL under black-box attacks. Following previous literature \cite{aleks2017deep,DBLP:journals/corr/PapernotMG16}, we apply transfer-based black-box attacks which generate adversarial samples using a victim model and feed them to the target models. In this paper, we take a ResNet-110 network as the victim model. The transferred adversarial samples are generated through FGSM, MIM and PGD attacks. The adversarial accuracy of different architectures is compared after they are fed with these transferred adversarial samples, as shown in Table \ref{table:black_box_NAS}. \textcolor{black}{Compared with other standard NAS algorithms, RACL achieves the highest robust accuracy in all the scenarios under these transfer-based attacks, which highlights the adversarial robustness of the proposed algorithm against transfer-based black-box attacks.}

\noindent\textbf{Transferability Test on CIFAR-10 under PGD Attack}\quad Following \cite{guo2019nas}, we further conduct the transferability test on CIFAR-10. We use different NAS algorithms as source models to generate adversarial samples through 10-iteration PGD attack and feed them to other target models as cross black-box attacks. The results are shown in Table \ref{table:black_box_transfer}. Each row denotes the robust accuracy of different target models under the black-box attack from the same source model. Correspondingly, each column denotes the robustness of a target model under attack from different source models.
\textcolor{black}{Comparing each row, RACL achieves the best accuracy under the attacks from different source models, which indicates that although these architectures are searched within the same search space, they show different robustness under attacks. The large gap between RACL and other baselines also highlights the superiority of RACL under black-box settings. Furthermore, through comparing the transferability between each model pair, RACL tends to generate stronger adversarial samples. \textit{E}.\textit{g}., RACL $\to$ AmoebaNet achieves the successful attack success rate of $35.52\%$ and AmoebaNet $\to$ ours achieves the successful attack success rate of $29.59\%$. Taking NASNet, DARTS and PC-DARTS as target models, RACL generates the adversarial samples which achieve the highest attack success rate except for the white-box attack.}

\begin{table}[t]
		\caption{\textcolor{black}{Transferability test on CIFAR-10 among differnet models using PGD attack. The best results in each row are in bold. Underline denotes the white-box robustness.}} \label{table:black_box_transfer}
	\begin{center}
		\scalebox{0.85}{
			\color{black}\begin{tabular}{c|c|c|c|c|c}
				\hline
				\hline
				\diagbox{Source}{Target} & AmoebaNet & NasNet & DARTS & PC-DARTS & ours\\ \hline
				AmoebaNet  & \underline{53.69} & 66.79 & 64.39 & 66.50 & \bf{67.21}\\
				NasNet     & 64.77 & \underline{53.35} & 64.31 & 65.62 & \bf{66.22}\\
				DARTS      & 65.03 & 66.91 & \underline{54.04} & 66.74 & \bf{67.04}\\
				PC-DARTS   & 64.37 & 65.45 & 63.91 & \underline{53.01} & \bf{66.23}\\
				\hline
				RACL(ours) & 64.49 & \bf66.24 & 64.06 & 65.90 & \underline{55.68}\\
				\hline
				\hline
			\end{tabular}
		}
	\end{center}
\end{table}

\noindent\textbf{Transferability Test on CIFAR-10 under RFGSM Attack}\quad
\begin{table}[b]
	\caption{\textcolor{black}{Transferability Test on CIFAR-10 among different models under RFGSM Attack. The best results in each row are in bold. Underline denotes the white-box robustness.}}
	\label{table:Transferability}
	\centering
	\scalebox{0.85}{
		\color{black}\begin{tabular}{c|c|c|c|c|c}
			\hline
			\hline
			\diagbox{Source}{Target} & AmoebaNet & NasNet & DARTS & PC-DARTS & ours\\ \hline
			AmoebaNet  & \underline{76.68} & 80.97 & 77.13 & 80.93 & \bf{81.22}\\
			NasNet     & 78.93 & \underline{78.52} & 77.18 & 80.85 & \bf{81.09}\\
			DARTS      & 78.81 & 80.92 & \underline{75.26} & 80.82 & \bf{81.16}\\
			PC-DARTS   & 78.70 & 80.68 & 77.02 & \underline{78.43} & \bf{81.16}\\
			\hline
			RACL(ours)       & 78.78 & 80.73 & 77.18 & \bf{80.83} & \underline{78.83}\\
			\hline
			\hline
	\end{tabular}}
\end{table}
\textcolor{black}{Furthermore, we provide additional robustness evaluation of RACL through transferability test on CIFAR-10 under RFGSM attack \cite{2018ensemble}. The detailed results are shown in Table \ref{table:Transferability}. The underline denotes the adversarial accuracy under white-box RFGSM attack where the total perturbation is set to $8/255$. Comparing the accuracy on diagonal, RACL achieves the best white-box performance under RFGSM attack. Each row denotes the robustness of different target models under the black-box attack from the same source model. Comparing each column, RACL shows strong adversarial transferability as the source model. Comparing each row, RACL achieves better black-box adversarial accuracy in all the scenarios as shown in Table \ref{table:Transferability} with bold. \textit{E}.\textit{g}., as shown in the fourth row, PC-DARTS $\to$ RACL achieves the successful attack success rate of $18.84\%$, PC-DARTS $\to$ AmoebaNet of $21.30\%$, PC-DARTS $\to$ NASNet of $19.32\%$ and PC-DARTS $\to$ DARTS of $22.98\%$. Similarly, RACL achieves strong adversarial transferability with RFGSM attack. Thus, RACL shows superior adversarial robustness against different transferred-based attacks, which demonstrates the effectiveness of our algorithm.}

\begin{figure*}[t]
	\begin{center}
		\includegraphics[width=1.0\linewidth]{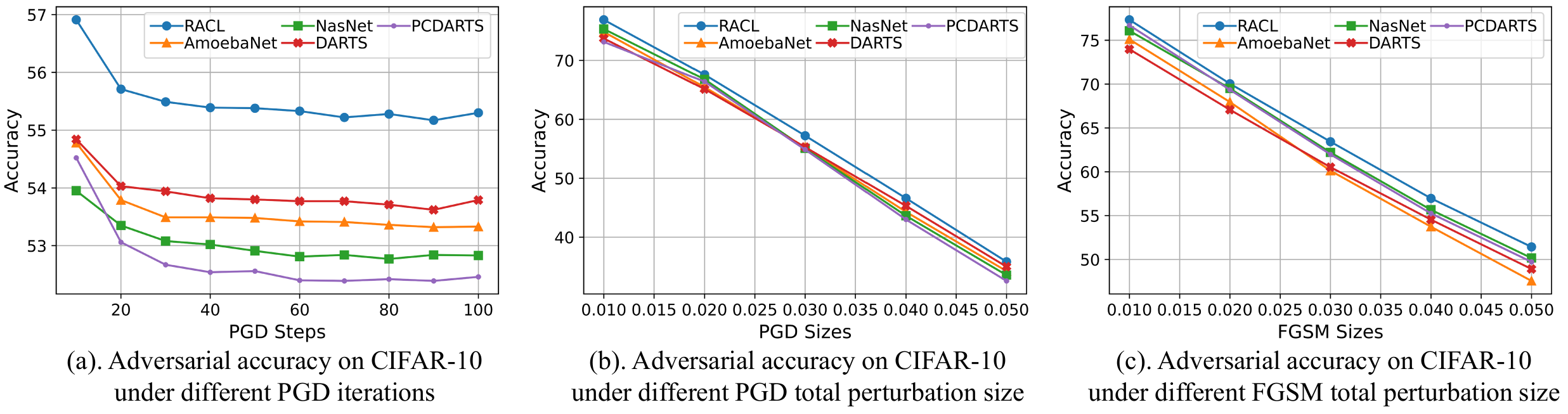}
	\end{center}
	\caption{\textcolor{black}{Robustness evaluation under different perturbation sizes and attack iterations.}}
	\label{fig:attack_study}
\end{figure*}

\subsection{Robustness under Various Perturbation Size and Attack Iterations}
\noindent\textbf{Robustness under Increasing Attack Iterations}\quad
\textcolor{black}{We further conduct experiments with different white-box attack parameters, including the size of perturbation and the number of iterations. Following \cite{guo2019nas}, we strengthen the adversarial attack through boosting the attack iterations to 100 for PGD attack with a step size of $2/255$. The comparison with other baselines is shown in Figure \ref{fig:attack_study} (a) where RACL consistently achieves the best accuracy for different PGD iterations. Furthermore, RACL shows relatively stronger defence capability against PGD attacks with more iterations. For example, NasNet achieves $53.35\%$ under PGD$^{20}$ and $52.83\%$ under PGD$^{100}$ with a gap of $0.52\%$ while RACL achieves $55.68\%$ under PGD$^{20}$ and $55.32\%$ under PGD$^{100}$ with a gap of $0.36\%$, which demonstrates that RACL can better remain the robustness after more attack iterations. Compared to RobNet family with the same search space on PGD$^{100}$ \cite{guo2019nas}, RACL achieved better performance with fewer parameters than RobNet-small, RobNet-medium, RobNet-large, and RobNet-free but $7.25\%$, $6.36\%$, $6.08\%$, and $2.75\%$ accuracy improvement respectively, which also shows the efficiency of RACL.}

\noindent\textbf{Robustness under Increasing Perturbation Size}\quad
\textcolor{black}{Besides attack iterations, we evaluate the adversarial robustness under different perturbation budgets. As shown in Figure \ref{fig:attack_study} (b, c), the total perturbation size ranges from 0.01 to 0.05 for both PGD and FGSM attacks. Our proposed RACL algorithm always performs better than other baselines under different perturbation budgets, which illustrates that RACL can provide a stronger defence against various adversarial attacks. Similarly, our advantage becomes more obvious when the attack was allowed with a larger total perturbation size. \textit{E}.\textit{g}., comparing NasNet with RACL on the PGD$_{0.01}$ and PGD$_{0.05}$, the gap increases $0.71\%$ when the attack size grew ($75.33\% \to 76.89\%$ on PGD$_{0.01}$ and $33.57\% \to 35.84\%$ on PGD$_{0.05}$); comparing AmoebaNet with RACL on the FGSM$_{0.01}$ and FGSM$_{0.05}$, the gap increases $1.65\%$ ($75.10\% \to 77.33\%$ on FGSM$_{0.01}$ and $47.55\% \to 51.43\%$ on FGSM$_{0.05}$), which highlights the adversarial robustness of RACL within a wider perturbation space for various attacks.}

\begin{table}[t]
		\caption{\textcolor{black}{Multiple runs of searched cells with adversarial training. The mean of clean or adversarial accuracy is reported with its error bar.}} \label{table:variance_trial}
	\begin{center}
	\setlength{\tabcolsep}{3pt} 
		\scalebox{0.8}{
			\color{black}\begin{tabular}{c|c|ccccc}
				Models & Clean & FGSM & PGD$^{20}$ & MIM & CW & AA\\
				\hline
				\hline
				AmoebaNet & 81.86$\pm$0.22 & 59.19$\pm$0.32 & 53.43$\pm$0.45 & 57.07$\pm$0.49 & 78.33$\pm$0.63 & 47.64$\pm$0.37\\
				NasNet & \bf84.05$\pm$0.64 & 59.49$\pm$0.53 & 53.37$\pm$0.70 & 58.16$\pm$0.55 & 79.57$\pm$0.74 & 48.34$\pm$0.46\\
				DARTS & 80.84$\pm$0.88 & 59.49$\pm$0.53 & 53.82$\pm$0.59 & 57.32$\pm$0.45 & 77.34$\pm$0.99 & 48.31$\pm$0.38\\
				PC-DARTS & 84.01$\pm$0.68 & 60.85$\pm$0.26 & 53.30$\pm$0.51 & 58.17$\pm$0.46 & 79.93$\pm$0.56 & 47.56$\pm$0.50\\
				\hline
				RACL & 83.98$\pm$0.22 & \bf62.48$\pm$0.10 & \bf55.50$\pm$0.38 & \bf60.01$\pm$0.39 & \bf80.11$\pm$0.39 & \bf50.14$\pm$0.33\\
				\hline
				\hline
			\end{tabular}
		}
	\end{center}
\end{table}
\begin{figure*}[t]
	\begin{center}
		\includegraphics[width=1.0\linewidth]{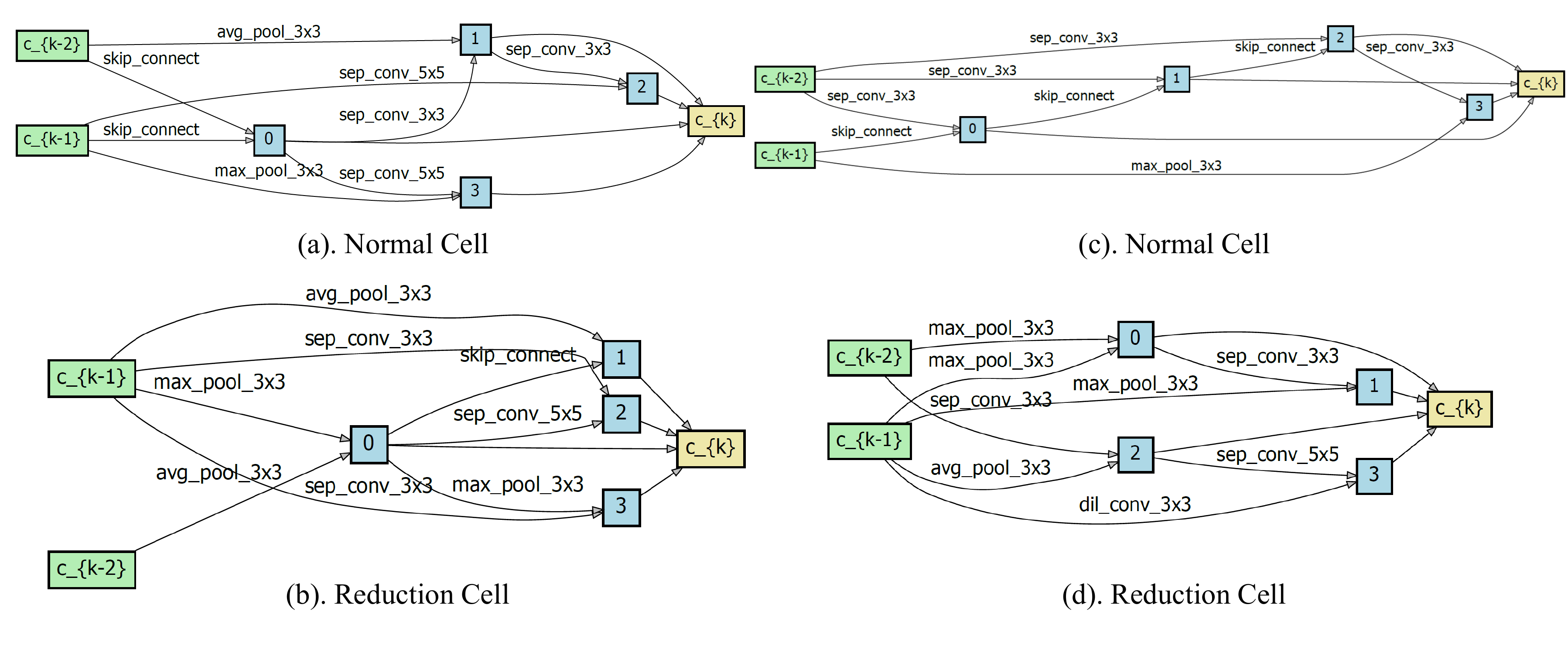}
	\end{center}
	\caption{\textcolor{black}{The visualization of cells searched by RACL through multiple runs.}}
	\label{fig:Cells}
\end{figure*}

\begin{table}[t]
		\caption{\textcolor{black}{Ablation Analysis of RACL with respect to confidence learning, $\rho$, and $\eta$.}} \label{table:Ablation_Analysis}
	\begin{center}
		\scalebox{0.92}{
			\color{black}\begin{tabular}{c|c|ccccc}
				Setting & Clean & FGSM & PGD$^{20}$ & MIM & CW & AA\\
				\hline
				Random Search & 81.55 & 58.73 & 51.36 & 55.72 & 76.63 & 46.82\\
                w/o Gradient Norm & 82.54 & 60.33 & 53.87 & 58.04 & 78.94 & 48.25\\
				w/o CL & \bf84.35 & 61.64 & 54.58 & 58.68 & 78.72 & 48.89\\
				\hline
				$\eta=0.9$, $\rho=0.01$ & 83.06 & 61.60 & 55.66 & 59.34 & 79.51 & 48.98\\
				$\eta=0.7$, $\rho=0.001$ & 84.30 & 61.58 & 55.00 & 59.14 & 80.36 & 49.03\\
				\hline
				$\eta=0.9$, $\rho=0.001$  & 84.04 & \bf62.07 & \bf55.68 & \bf60.00 & \bf80.90 & \bf50.07\\
				\hline
				\hline
			\end{tabular}
		}
	\end{center}
\end{table}

\subsection{Potential Pattern and Variance of RACL} \label{sec: variance}
\noindent\textbf{Visualization of Searched Cells}\quad
\textcolor{black}{Besides the cells visualized in Fig. \ref{fig:Cell}, we run RACL several times to explore the potential patterns RACL tended to discover and provided more insights into the searched robust neural architectures. More searched architectures are given in Fig. \ref{fig:Cells}. Together with the searched cells in Fig. \ref{fig:Cell}, we showed that there exist some potential patterns which RACL prefers. 
There always exists a ResNet-like pattern in the searched normal cells. For example, the input of each node tends to be a combination of skip connection and another operation with trainable parameters, such as the Node $0$, $1$, $2$ in Fig. \ref{fig:Cell} (a) and Fig . \ref{fig:Cells} (c). Besides the ResNet-like pattern in the normal cells, RACL tends to select pooling layers such as $3\times3$ max pooling instead of skip connection in the reduction cells, as shown in Figure \ref{fig:Cell} (b) and Figure \ref{fig:Cells} (b), (d). Overall, the searched cells of RACL look like a tuned version of ResNet.
}

\noindent\textbf{Robustness Stability of Searched Cells}\quad
\textcolor{black}{To further demonstrate the effectiveness of RACL, we report the error bar of RACL as well as other baselines through multiple runs to evaluate the robustness stability of searched cells. Following previous NAS work \cite{DBLP:journals/corr/abs-1806-09055}, we retrain all the baselines for multiple times and report the performance of neural architectures searched by different NAS algorithms and RACL (out of 5 runs). The detailed results are shown in Table \ref{table:variance_trial}. For each algorithm, we report the average clean and robust accuracy with the standard error. Comparing each column, RACL consistently achieves the best average robust accuracy under various attacks after multiple runs. For example, RACL achieves an average accuracy of $50.14\%$ under Auto Attack and $55.50\%$ under PGD$^{20}$, which is around $2\%$ higher than other baselines. For the error bar, RACL has smaller fluctuation in almost all the scenarios among different NAS algorithms, which demonstrates that RACL can discover robust neural architectures with better robustness and stability.}

\subsection{Ablation Analysis}
\textcolor{black}{In this section, we conduct ablation studies on the hyperparameters of RACL algorithm as well as confidence learning. The ablation study results are shown in Table \ref{table:Ablation_Analysis}. We first apply the random search algorithm within the pre-defined search space to rule out the possibility that the major improvement comes from the search space. We randomly sampled 10 models and selected the best one for comparison. We then remove the confidence learning and apply the constraint in Eq. \ref{eq:lipschitz_full_eq} to evaluate the effectiveness of confidence learning. Similarly, we remove the gradient norm constraint in Eq. \ref{eq:constraint_objective} to evaluate the effectiveness of lower bound constraint.
Comparing the first and other rows, the random search algorithm cannot achieve competitive results within a pre-defined search space, which demonstrates the necessity of discovering robust neural architectures.
Comparing the second and last row, the searched architecture without confidence learning tends to have a relatively higher natural accuracy. On the contrary, our proposed RACL achieves a relatively large increment in adversarial accuracy with confidence learning, which highlighted the importance of proposed confident architecture sampling.
We then investigated the influence of hyperparameter $\rho$ and reported the performance of searched robust cell on CIFAR-10 under different values of $\rho$. Through comparison, $\rho$ with a large value could hurt the classification performance on clean images. On the contrary, $\rho$ with a small value reduces the influence of Lipschitz constraint and results in inferior adversarial accuracy. The influence of confidence hyperparameter $\eta$ is also investigated. From Eq. \ref{eq:constraint_objective}, $\eta$ controls the balance between the mean and variance of Lipschitz constant $\overline{\lambda_\mathcal{F}}$. Through cross-validation, $\eta$ is set to $0.9$ to obtain the best adversarial accuracy.}

\section{Conclusion}
In this paper, we propose to tackle the vulnerability of neural networks by incorporating NAS frameworks. Through sampling architecture parameters from trainable log-normal distributions, we show that the approximated Lipschitz constant of the entire network can be formulated as a univariate log-normal distribution, which enables the proposed algorithm, Robust Architecture with Confidence Learning to form confidence learning of architecture parameters on the robustness through a Lipschitz constraint. Thorough experiments demonstrate the influence of architecture on adversarial robustness and the effectiveness of RACL under various attacks on different datasets.


%





\ifCLASSOPTIONcaptionsoff
  \newpage
\fi



%

\bibliographystyle{IEEEtran}
\bibliography{IEEEabrv,main}

\end{document}